\documentclass[
]{ceurart}

\usepackage{minted}
\setminted{breaklines=true}
\usepackage{enumitem}
\usepackage{algorithm}
\usepackage{algpseudocode}
\usepackage{pifont}
\newcommand{\cmark}{\ding{51}}
\newcommand{\xmark}{\ding{55}}
\usepackage{wrapfig}

\usepackage{tikz}
\usetikzlibrary{decorations.pathmorphing,shapes,positioning,arrows.meta,arrows,automata,positioning,calc,shadows}

\usepackage{xargs} 
\let\todo\relax
\usepackage[colorinlistoftodos]{todonotes}
\setuptodonotes{size=tiny}

\newcommandx{\complete}[2][1=]{\todo[linecolor=orange,backgroundcolor=orange!25,bordercolor=orange,#1]{Complete: #2}}
\newcommandx{\tocite}[2][1=]{\todo[linecolor=pink,backgroundcolor=pink!25,bordercolor=pink,#1]{Cite: #2}}
\newcommandx{\unsure}[2][1=]{\todo[linecolor=blue,backgroundcolor=blue!5,bordercolor=blue,#1]{Unsure: #2}}
\newcommandx{\change}[2][1=]{\todo[linecolor=red,backgroundcolor=red!25,bordercolor=red,#1]{Change: #2}}
\newcommandx{\info}[2][1=]{\todo[linecolor=olive,backgroundcolor=olive!25,bordercolor=olive,#1]{Info: #2}}
\newcommandx{\improvement}[2][1=]{\todo[linecolor=purple,backgroundcolor=purple!25,bordercolor=purple,#1]{Improve: #2}}
\newcommandx{\cut}[2][1=]{\todo[linecolor=yellow,backgroundcolor=yellow!25,bordercolor=yellow,#1]{Potential Cut: #2}}

\newcommand{\attacks}[0]{\rightsquigarrow}

\newcommand{\args}[0]{\mathcal{A}}

\usepackage{bm}

\newcommand{\x}[1]{x_{#1}}
\newcommand{\y}[1]{y_{#1}}

\newcommand{\argalpha}{a}
\newcommand{\argbeta}{b}
\newcommand{\arggamma}{c}

\newcommand{\xalpha}{\x{\argalpha}}
\newcommand{\xbeta}{\x{\argbeta}}
\newcommand{\xgamma}{\x{\arggamma}}

\newcommand{\yalpha}{\y{\argalpha}}

\newcommand{\ygamma}{\y{\arggamma}}

\newcommand{\case}[2]{(\x{#1}, \y{#2})}
\newcommand{\casealpha}{\case{\argalpha}{\argalpha}}
\newcommand{\casebeta}{\case{\argbeta}{\argbeta}}
\newcommand{\casegamma}{\case{\arggamma}{\arggamma}}

\newcommand{\casedefault}{(\x{\delta}, \delta)}

\newcommand{\argnew}{N}
\newcommand{\casenew}{\case{\argnew}{?}}
\newcommand{\xnew}{\x{\argnew}}

\newcommand{\af}[2][\empty]{$AF_{#1}(D_{cb}, {#2})$}

\usepackage{amsmath}
\usepackage{amsfonts}
\usepackage{amssymb}
\usepackage{bm}
\usepackage{amsthm}
\numberwithin{theorem}{section} % important bit

\numberwithin{corollary}{section}

\numberwithin{lemma}{section}

\numberwithin{proposition}{section} 

\newtheorem{definition}{Definition}
\numberwithin{definition}{section}

\numberwithin{conjecture}{section}

\numberwithin{claim}{section}

\numberwithin{remark}{section}

\numberwithin{example}{section}

\numberwithin{hypothesis}{section}

\numberwithin{property}{section}

\numberwithin{assumption}{section}

\usepackage{amsmath,amsfonts,bm}

\def\eqref#1{equation~\ref{#1}}
\def\1{\bm{1}}

\def\mA{{\bm{A}}}

\DeclareMathAlphabet{\mathsfittwo}{\encodingdefault}{\sfdefault}{m}{sl}
\SetMathAlphabet{\mathsfittwo}{bold}{\encodingdefault}{\sfdefault}{bx}{n}

\begin{document}

\copyrightyear{2025}
\copyrightclause{Copyright for this paper by its authors.
  Use permitted under Creative Commons License Attribution 4.0
  International (CC BY 4.0).}

\conference{ArgXAI-25: 3rd International Workshop on Argumentation for eXplainable AI}

\title{Object-Centric Case-Based Reasoning via Argumentation}

\author[1]{Gabriel de Olim Gaul}[%
email=gabriel.de-olim-gaul21@imperial.ac.uk
]

\author[1]{Adam Gould}[%
orcid=0009-0008-0972-7501,
email=adam.gould19@imperial.ac.uk
]
\cormark[1]

\author[1]{Avinash Kori}[%
orcid=0000-0002-5878-3584,
email=a.kori21@imperial.ac.uk
]

\author[1]{Francesca Toni}[%
orcid=0000-0001-8194-1459,
email=f.toni@imperial.ac.uk
]

\address[1]{Department of Computing, Imperial College London, United Kingdom}

\begin{abstract}
We introduce Slot Attention Argumentation for Case-Based Reasoning (SAA-CBR), a novel neuro-symbolic pipeline for image classification that integrates object-centric learning via a neural Slot Attention (SA) component with symbolic reasoning conducted by Abstract Argumentation for Case-Based Reasoning (AA-CBR). We explore novel integrations of AA-CBR with the neural component, including feature combination strategies, casebase reduction via representative samples, novel count-based partial orders, a One-Vs-Rest strategy for extending AA-CBR to multi-class classification, and an application of Supported AA-CBR, a bipolar variant of AA-CBR. We demonstrate that SAA-CBR is an effective classifier on the CLEVR-Hans datasets, showing competitive performance against baseline models. 
\end{abstract}


\maketitle

%\todoin{
%
%\begin{itemize}
%    \itemdone Running and/or simplified Example
%    \itemdone More detail on explainability
%    \itemdone More discussion on the results - Why is the variance of SAA-CBR relatively high compared to the baselines, for example? 
%    \itemdone [page 5, at the end of the Notations paragraph] The 'weakly supervised fashion' in which the image is processed refers to noisier/limited/imprecise labels? This is unclear. More details are needed.
%    \itemdone fix typos:
%    \begin{itemize}
%        \itemdone -"... datasets[18]." => seems to missing a space
%        \itemdone - [page 1, Introduction, 15th line from the top] 'debate based' -> 'debate-based';
%        \itemdone - [page 1, Introduction, 17th line from the top] 'clearly' -> 'clear';
%        \itemdone - [page 3, 4th line from the top] 'extractor is converts' -> 'extractor converts';
%        \itemdone - [page 5, Neural Component, 10th line from the top] 'using week supervision' -> 'using weak supervision';
%        \itemdone - [page 6, 2nd line from the top] 'e.g' -> 'e.g.,';
%        \itemdone - [page 6, Multi-Class Classification, 4th line from the top] 'selected to' -> 'decided to' (?);
%        \itemdone - [page 7, Instantiating AA-CBR, 6th line from the bottom] There should be a colon rather than a comma after fi, isn't it?
%    \end{itemize}
%\end{itemize}

%}

\section{Introduction}

Over the past decade, deep learning models have become powerful tools across a wide range of domains. Despite their impressive performance, these models are often regarded as black boxes due to their lack of interpretability. Meanwhile, there is growing consensus among researchers, ethicists, policymakers, and the broader public about the critical importance of explainability — especially in high-stakes applications such as medicine, healthcare, and autonomous driving \cite{doshi2017role, kroll2015accountable}.
Interpreting the decisions made by deep learning classifiers not only highlights their internal mechanisms but also helps identify potential biases \cite{kim2018interpretability}, thereby deepening our understanding of the underlying data-generating process \cite{narayanaswamy2020scientific}. Many techniques have been developed to enhance interpretability, including feature attribution methods \cite{ribeiro2016should}, network dissection approaches \cite{bau2017network}, mechanistic analyses of neural networks \cite{olah2017feature, olah2020zoom}, and causal or counterfactual explanations \cite{sauer2021counterfactual}. Explainability methods are generally categorised as either ante-hoc or post-hoc explanations \cite{lipton2018mythos}, with the majority falling into the post-hoc category.

However, most of these approaches fail to align with the concept-based reasoning that humans naturally employ \cite{armstrong1983some}. Some recent methods aim to bridge this gap-\cite{ghorbani2019towards} highlights the presence of conceptual representations, while \cite{kori2020abstracting} explores their interactions to produce explanations. To further this direction, debate-based explanation generation was introduced in \cite{visualdebtes24, FAX_arXiv}, where the rationale for a prediction emerges from simulated agent dialogues. Nonetheless, these methods often fall short in producing clear human-interpretable concepts.

Neuro-symbolic methods have increasingly focused on addressing this particular challenge by integrating neural networks with symbolic, logic-based reasoning frameworks \cite{neuro-symbolic-ai-3rd-wave, RefWorks:RefID:1-proietti2023roadmap}. These hybrid models combine the representational power of neural networks, which excel at learning complex data patterns, with the structured reasoning capabilities of symbolic systems, leading to more interpretable outputs \cite{RefWorks:RefID:2-stammer2021concept:, jacob2025objectcentric}.
Building on this foundation, we propose a novel and scalable neuro-symbolic framework that leverages object-centric learning—specifically Slot Attention (SA)~\cite{locatello2020object} as the neural backbone for unsupervised, human-understandable concept discovery. 
Here, the neural component extracts a set of object representations, which are then classified to derive object-level attributes. These attributes generate arguments within a symbolic argumentation model to predict outcomes for input images. Unlike prior work that combines SA with argumentation \cite{jacob2025objectcentric}, our approach is the first to incorporate the Abstract Argumentation for Case-Based Reasoning (AA-CBR) framework \cite{aa-cbr}, which leverages prior argumentation cases to resolve new instances—offering enhanced scalability.

\paragraph{Contributions:} Our key contributions are as follows. (1) We propose Slot Attention Argumentation for Case-Based Reasoning (SAA-CBR), a novel neuro-argumentative framework for image classification, and evaluate its performance on the CLEVR-Hans datasets~\cite{RefWorks:RefID:2-stammer2021concept:}, demonstrating competitive results with state-of-the-art models.
%(2) We introduce Supported AA-CBR, a Bipolar Argumentation Framework that extends AA-CBR with support relations, and propose two definitions for supported attacks. Despite its expressive power, we demonstrate that Supported AA-CBR does not improve classification performance within the SAA-CBR pipeline;
(2) We explore novel integrations of AA-CBR into the SAA-CBR pipeline. These include using a One-Vs-Rest approach to apply the model in a multi-class classification setting; experimenting with Supported AA-CBR~\cite{supported-aacbr}, a novel variant that introduces support relations in the casebase; using case clustering and uncertainty filtering to reduce the size of the casebase; exploring novel methods for feature combination to overcome AA-CBR's inherent lack of feature weighting.
(3) 
%We explore approximate AA-CBR, a variant that drops the minimality constraint to yield a flat framework. We show it to be both data-efficient and scalable. 
Additionally, we propose novel counting-based partial orders and characterisations for AA-CBR, enabling object-type counting within scenes as part of the proposed pipeline.

\section{Related Work}

% \todoin{
% \begin{itemize}
%     \item Abdul's paper (https://doi.org/10.48550/arXiv.2506.14577)
%     \item NAL paper (https://ceur-ws.org/Vol-3432/paper1.pdf)
%     \begin{itemize}
%         \item DEAR/ANNA pipeline
%         \item ProtoArgNet?
%     \end{itemize}
%     \item Gradual AA-CBR (https://doi.org/10.48550/arXiv.2505.15742)
%     \item Neuro-Symbolic Concept Learner
%     \item Slot-Attention paper
%     \item ArgLLMs? 
%     \end{itemize}
% }

\paragraph{Object-centric representation learning.} 
The primary goal of Object-Centric Learning (OCL) is to achieve spatial disentanglement—that is, to learn independent representations for each object present in an image. Early approaches to OCL~\cite{burgess2019monet, engelcke2019genesis, greff2019multi} predominantly leveraged (iterative) variational inference procedures~\cite{marino2018iterative, van2020investigating, lin2020space}, typically built upon the Variational Autoencoder (VAE) framework~\cite{kingma2013auto}.
More recent advances in the field have shifted toward using an iterative attention mechanism known as Slot Attention (SA)~\cite{locatello2020object} to fulfil the OCL objective. A growing body of follow-up work~\cite{engelcke2021genesis, singh2021illiterate, wang2023slot, singh2022neural, emami2022slot, kori2023grounded, kori2024identifiable, seitzer2022bridging} has since extended the Slot Attention framework, making it applicable to a range of real-world settings.
In this work, we adopt the iterative Slot Attention approach to spatially disentangle image content and leverage the resulting object-centric representations for downstream reasoning tasks.

\paragraph{Neuro-symbolic learning.}
\cite{RefWorks:RefID:1-proietti2023roadmap} provides an overview of several approaches that integrate logic-based learning with neural models, including for image classification. One such method, NeSyFold~\citep{RefWorks:RefID:26-padalkar2023nesyfold:}, employs the rule-based learning algorithm FOLD-SE-M to transform binarized kernels from a trained convolutional neural network into a set of Answer Set Programming (ASP) rules~\citep{GeL91} with abstract predicates. It then applies semantic labeling to associate these predicates with human-interpretable concepts, thereby producing global explanations for image classifications.
Another method, Embed2Sym~\citep{RefWorks:RefID:40-aspis2022embed2sym}, combines clustered neural embeddings with a symbolic reasoner, encoded with predefined rules, to generate explainable predictions. Additional approaches incorporate argumentation-based reasoning for image classification. For example, \cite{ECAI24imagesarg} classifies images through argumentative debates derived from neural encoders, while \cite{visualdebtes24, FAX_arXiv} generate argumentative explanations based on quantized image features.
However, most of these approaches either depend on densely labeled data or lack the ability to produce truly human-understandable concepts and explicit symbolic reasoning. This limitation motivated the development of the Neuro-Symbolic Concept Learner (NS-CL)~\citep{RefWorks:RefID:2-stammer2021concept:}, which leverages object-centric learning via Slot Attention to identify distinct objects in images, while using set transformers in its reasoning module to support explanation generation.
This line of work was further extended to incorporate fully symbolic reasoning layer in Object-Centric Neuro-Argumentative Learning (OC-NAL)~\citep{jacob2025objectcentric}, which enables more complex argumentative reasoning over the slot-based object representations extracted from visual input.
In this work, inspired by OC-NAL, we enable case-based reasoning on the extracted object representations, providing access to the model's reasoning for making a certain decision.  

\paragraph{Argumentative Case-Based Reasoning} Computational argumentation has been used to support case-based reasoning approaches, in which previously observed cases are cited as evidence for/against a classification. \textit{Abstract Argumentation for Case-Based Reasoning (AA-CBR)}~\cite{aa-cbr} is one such approach wherein data points are characterised as \textit{cases} that argue about the outcome of an unlabelled case. However, as a purely symbolic method, this method cannot easily be deployed with complex data types, such as images, and cannot automatically select features in the data. The introduction of user-specified feature preferences~\cite{preference-based-aacbr} to AA-CBR was proposed to aid feature selection, but this does not solve the issue of more complex data types. The \textit{Data-Empowered Argumentation (DEAr)}~\cite{DEAr} was proposed, in which a characterisation extractor converts complex data into symbols that can be reasoned over with AA-CBR. However, DEAr has not been %extensively 
tested with a neural component extracting features from images. Our proposed approach uses this pipeline methodology with an SA module as the feature extractor. This is unlike Gradual AA-CBR~\cite{Gradual-AACBR}, which attempts to learn the argumentation debate end-to-end with the feature extractor but has yet to be applied to more complex data types like images.

\section{Background}

\subsection{Slot-Attention}
The Slot Attention (SA) module~\citep{locatello2020object} is an architectural mechanism that interfaces with perceptual spatial representations, as estimated by a convolutional neural network (CNN) encoder—to produce a set of task-specific abstract representations referred to as slots. 
These slots dynamically bind to arbitrary objects in the input through a competitive, attention-based refinement process that unfolds over multiple iterations.
Slot attention takes a set of $N$ feature embeddings $\mathbf{z} = \phi(\mathbf{x}) \in \mathbb{R}^{N \times d_z}$, as input, where $d_{z}$ is the dimensionality of $\mathbf{z}$, $\mathbf{x}$ corresponds to an observed image, with CNN encoder $\phi$, and applies an iterative attention to produce $K$ object representations called slots $\mathbf{s} \in \mathbb{R}^{K \times d_s}(d_s \leq d_z)$, where $d_{s}$ is the dimensionality of each slot.
For this the input features $\mathbf{z}$ are projected with a linear layer resulting in a \emph{query, key}, and \emph{value} representations, denoted by $\mathbf{q}, \mathbf{k}$ and $\mathbf{v}$, respectively.
To simplify 
%our exposition later on, 
let $f$ and $g$ be shorthand notation for the \textit{slot update} and \textit{attention} functions respectively:

\begin{align}
    f(\mathbf{A},\mathbf{v}) = \mathbf{A}^T\mathbf{v}, \quad \mathbf{A}_{ij} = 
    \frac{g(\mathbf{q}, \mathbf{k})_{ij}}{\sum_{l=1}^K g(\mathbf{q}, \mathbf{k})_{lj}}
    \quad && \mathrm{and}  &&
    g(\mathbf{q},\mathbf{k}) = \frac{e^{\mathbf{M}_{ij}}}{\sum_{l=1}^N e^{\mathbf{M}_{il}}}, \quad 
    % \mathrm{with} \quad 
    \mathbf{M} =\frac{\mathbf{k}\mathbf{q}^T}{\sqrt{d_s}},
    \label{eqn:fgequation}
\end{align}

where $\mA \in \mathbb{R}^{N \times K}$ reflects the cross-attention matrix capturing the competition across $K$ slots to attend to a $N$ image tokens. 
Unlike self-attention~\cite{vaswani2017attention}, the queries in slot attention are a function of the slots, initialised by sampling from a normal distribution $\mathcal{N}(\mathbf{s}; \mathbf{\mu}, \mathbf{\sigma})$, i.e.,
$\mathbf{s} \sim \mathcal{N}(\mathbf{s}; \mathbf{\mu}, \mathbf{\sigma}) \in \mathbb{R}^{K\times d_s}$,
and are iteratively refined over $T$ attention iterations~\citep{locatello2020object}. 
Methodologically, the slots are randomly initialized at $t=0$. The queries at iteration $t$ are given by $\hat{\mathbf{q}}^t$ and are generated as a linear projection of current state slots $\mathbf{s}^t$, and the slot update process can be summarized as: $\mathbf{s}^{t+1} = f(g(\hat{\mathbf{q}}^t, \mathbf{k}), \mathbf{v})$. Lastly, a Gated Recurrent Unit (GRU) is applied to the slot representations $\mathbf{s}^{t+1}$ at the end of each SA iteration, followed by a generic MLP skip connection.

\subsection{Abstract Argumentation for Case-Based Reasoning}
An abstract \emph{argumentation framework (AF)}~\citep{DUNG-aa} is a pair $\langle \args, \attacks \rangle$, where $\args$ is a set of arguments and $\attacks \subseteq \args \times \args$ is a binary relation defining \emph{attacks} between arguments. We can visualise an AF as a directed graph with arguments as nodes and attacks as edges.
A set of arguments $E \subseteq \args$ \emph{defends} an argument $\argbeta \in \args$ if for all $\argalpha \attacks \argbeta$ there exists $\arggamma \in E$ such that $\arggamma \attacks \argalpha$. To determine the set of \textit{acceptable} arguments, we apply the \textit{grounded} semantics. The \textit{grounded extension} $\mathbb{G}$ is computed as $\mathbb{G} = \bigcup_{i\geq0}G_{i},$ where $G_{0}$ is the set of unattacked arguments and $\forall i \geq 0, G_{i+1}$ is the set of all arguments that $G_{i}$ defends.

\emph{Abstract Argumentation for Case-Based Reasoning (AA-CBR)}~\citep{aa-cbr} is a machine learning approach for binary classification in which an argumentation framework is constructed from the training data. Each argument in the resulting AF is built from a single data point and is referred to as a \emph{case}.
A partial order defining a notion of \emph{exceptionality} between data points allows the AF to be constructed utilising defeasible reasoning, wherein each data point is an exception to those they attack. An unlabelled case, $N$, is added to the framework by attacking any cases considered \textit{irrelevant} to it. We then apply the grounded semantics to compute which arguments are accepted. If a \textit{default argument} - representing our expected outcome given no other information - is accepted, then its corresponding outcome is predicted for $N$; otherwise, the opposing class is predicted. Formally:

\begin{definition}
    \label{def:aa-cbr}
    Let $D_{cb} \subseteq X \times Y$ be a finite \emph{casebase} of labelled examples where $X$ is a set of \emph{characterisations} and $Y = \{\delta, \bar{\delta}\}$ is the set of possible outcomes. Each example is of the form $(x, y)$. Let $\casedefault$ be the \emph{default argument} with $\delta$ the \emph{default outcome}. Let $N$ be an \emph{unlabelled example} of the form $\casenew$ with $y_{?}$ an unknown outcome.  
    Let $\succcurlyeq$ and $\nsim$ be a partial order and binary relation defined over $X$ representing \textit{exceptionality} and \textit{irrelevance} respectively. The argumentation framework \af{\xnew} mined from $D_{cb}$ and $x_N$ is $\langle\args, \attacks\rangle$ in which:
    
    \begin{itemize}
        \item $\args = D_{cb} \cup \{\casedefault\} \cup \{N\}$
        \item for $\casealpha, \casebeta \in D_{cb} \cup \{(x_{\delta}, \delta)\}$, it holds that $\casealpha \attacks \casebeta$ iff
              \begin{enumerate}
                  \item $y_{\argalpha} \not = y_{\argbeta}$, and
                  % \item One of the following holds:  
                  % \begin{enumerate}
                      \item $\xalpha$ is more \emph{exceptional} than $\xbeta$ and there is \emph{minimal} difference between them, i.e.:
                    \begin{enumerate}
                      \item $\x{\argalpha} \succ \xbeta$ and \label{def:aa-cbr:exceptional}
                      \item $\not\exists \casegamma \in D_{cb} \cup \{(\x{\delta}, \delta)\}$ with $\ygamma = \yalpha$ and $\xalpha \succ \xgamma \succ \xbeta$; \label{def:aa-cbr:minimality} \hfill
                    \end{enumerate}
                    % \item or $\xalpha$ is equivalent to $\xbeta$: 
                  %   \begin{enumerate}
                  %     \item $\xalpha = \xbeta$; \label{def:aa-cbr:symmetric-attack}
                  % \end{enumerate}
                  \end{enumerate}
              % \end{enumerate}
        \item for $\casealpha \in D_{cb} \cup \{(\x{\delta}, {\delta})\}$, it holds that $N \attacks \casealpha$ iff $\xnew \nsim \xalpha.$
    
    \end{itemize}
    
    \noindent
     Finally, we have AA-CBR$(D_{cb}, \x{N}) = \delta$ if $\casedefault \in \mathbb{G}$ and $\bar \delta$ otherwise, where $\mathbb{G}$ is the grounded extension of \af{N}. 

\end{definition}

% A casebase $D$ is \emph{coherent} iff there are no two cases $\casealpha, \casebeta \in D$ such that $\xalpha = \xbeta$ and $y_{\argalpha} \neq y_{\argbeta}$, and it is \emph{incoherent} otherwise. 
In this work, we make use of \textit{regular} AA-CBR~\cite{monotonicity-and-noise-tolerance} in which the default characterisation, $x_{\delta}$, is the least element of the partial order $\succcurlyeq$ and irrelevance is the negation of exceptionality, that is~$\nsim = \not \succcurlyeq$.

% \todoin{

% \begin{itemize}
%     \itemtodo Spike
% \end{itemize}
% }

% As a result of enforcing minimality between cases (Condition \ref{def:aa-cbr:minimality}), not all arguments will have a path to the default case. These arguments, therefore, do not impact the final classification, even if they have features that may be relevant to the new case. These arguments are \emph{spikes}~\citep{monotonicity-and-noise-tolerance}. Formally:

% \begin{definition}
%     Let $\langle \args, \attacks \rangle$ = \af{\xnew} and $\argalpha \in \args$. $\argalpha$ is a \emph{spike} iff there is no path in $\langle \args, \attacks \rangle$ form $\argalpha$ to $\casedefault$.
% \end{definition}

% It is practical to define a notion of \emph{regular} AA-CBR~\citep{monotonicity-and-noise-tolerance} in which the choice of default case is such that $\x{\delta}$ is the least element of $X$ under the partial order $\succcurlyeq$ and a case $\argalpha$ is irrelevant to a new case $\argnew$ if $\argnew$ is not an exception to $\argalpha$. Formally:
% \unsure{probably do not need regular AA-CBR defined}

% \begin{definition} 
% \label{def:regular-aacbr}
% The AF mined from $D$ and $\xnew$ with default argument $\casedefault$ is \emph{regular} when:
% \begin{enumerate}
%     \item $\xalpha \nsim \xbeta$ iff $\xalpha \not \succcurlyeq \xbeta$, and
%     \item $\x{\delta}$ is the least element of $X$.
% \end{enumerate}
% \end{definition}

\section{Slot-Attention for Case-Based Reasoning}

\begin{figure}[t!]
    \centering
    \includegraphics[width=0.85\linewidth]{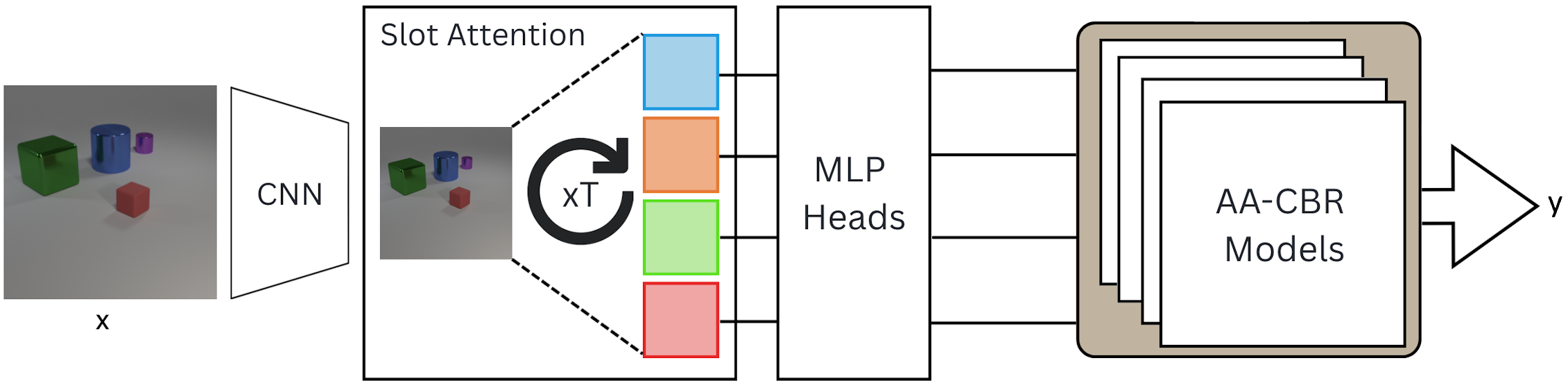}
    \caption{An overview of the Slot Attention Argumentation for Case-Based Reasoning (SAA-CBR) architecture.}
    \label{fig:saacbr_overview}
\end{figure}
% \todo{FT: should x be present in the figure? unclear what xT is... AG: {xT means `times T', i.e. the T slot attention iterations. The original paper slot attention paper includes xT in its diagrams}}

Figure~\ref{fig:saacbr_overview} showcases our proposed architecture, comprised of a neural-component governed by a slot-attention mechanism and a symbolic-component that reasons over the learned representations using AA-CBR. Effectively, the slot-attention mechanism acts as a characterisation extractor for AA-CBR. 

\paragraph{Notations.} Similar to OC-NAL~\cite{jacob2025objectcentric}, our architecture operates on an observed dataset $ D \subseteq \mathcal{X} \times \mathcal{Y} \times \mathcal{L} $ consisting of labelled images, where $ \mathcal{X} $ denotes the set of input images, $ \mathcal{L} = \{c_1, \dots, c_L\} $ represents a set of $L$ class labels, and $ \mathcal{Y} = \{0,1\}^{K \times (P + 1)} $ is metadata encoded as one-hot vectors. Here, $ K $ is the maximum number of objects (or slots) that can appear in an image, and $ P $ is the number of distinct properties each object may possess. An additional dimension is included to account for the absence of an object.
Importantly, the neural component of the architecture does not utilize the class labels in $ L $; instead, it processes the image data in $\mathbf{x}\sim\mathcal{X} $ in a \emph{weakly supervised} fashion. { This is weakly supervised because we are using object-level labels instead of pixel-wise dense labels. Additionally, semantics object segmentation is obtained in an unsupervised manner, while supervision is only used for classification.} Conversely, the symbolic reasoning component does not operate on the raw images but instead consumes an abstract representation generated by the neural component.

\subsection{Neural Component}

% \todo{is this different from Abdul's? I would add a commentary here as to the differences... It is same} 
The neural component is responsible for converting an image input $\mathbf{x} \in \mathcal{X}$ into a set of slot representations $\mathbf{s}$ that describe individual objects present in an image. 
These representations form the basis of the cases used by the argumentative reasoning component. 
To achieve this, we utilize the Slot Attention (SA) module, as described in the background section.
This component matches with the neural components described in \cite{jacob2025objectcentric}. 

The image $\mathbf{x}$ is first passed through a feature extractor $\phi$, yielding a feature map $\mathbf{z}$. 
The SA module then processes $\mathbf{z}$ in an iterative manner, producing a set of slot representations $\mathbf{s} = \{\mathbf{s}_1, \dots, \mathbf{s}_K\}$. 
Due to the permutation equivariant nature of SA, these slots are not inherently grounded with respect to object identity or properties. 
Therefore, we employ multi-layer perceptron (MLP) heads to classify the slot representations and derive object-level attributes (e.g., color, shape, size), using weak supervision from the available metadata $\mathcal{Y}$.

Our framework adopts a two-stage training approach, where the neural component is trained independently from the argumentative component. 
The object discovery mechanism within the neural component is optimized via self-supervision using a reconstruction objective. Specifically, we minimize the mean squared error (MSE) loss between the original image $\mathbf{x}$ and its reconstruction $\hat{\mathbf{x}}$. 
In parallel, the MLP heads are trained using a binary cross-entropy (BCE) loss between the predicted attributes $\hat{\mathbf{y}}$ and the ground-truth labels $\mathbf{y} \in \mathcal{Y}$, accounting for the multi-label nature of the classification problem.

As noted, the permutation equivariance of SA leads to an unknown slot order during inference. 
To address this, we apply the Hungarian Matching (HM) algorithm during training to find the optimal alignment $\tau$ between the predicted slots and ground-truth annotations before computing the BCE loss:

\begin{equation} \label{saacbr_loss}
    \mathcal{L} = \|\mathbf{x} - \hat{\mathbf{x}}\|_2^2 + \alpha \cdot \min_\tau \sum_{j=1}^P \mathrm{BCE}\left(\mathbf{y}_j, \hat{\mathbf{y}}_{\tau(j)}\right)
\end{equation}

where $\tau$ denotes the permutation provided by the HM algorithm, and $\alpha$ is a weighting hyperparameter. 
It is important to note that the HM step is used only during training for alignment with ground-truth annotations. During inference, this step is omitted. The final output of the neural component is a tensor containing the predicted attributes for each object in the input image.

\subsection{Argumentative Component}
%\todoin{
%\begin{itemize}
%    \itemdone One-Vs-Rest
%    \itemdone Thresholding
%    \itemdone K-Means Cluster
%    \itemdone Feature selection
%    \itemdone Supported AA-CBR X
%    \itemdone Proposed partial orders
%    \itemdone All possible model hyperparameters (e.g. choice of characterisation, default outcome class, confidence threshold, number of k-means centroids, use of supports or not) 
%\end{itemize}
%}

The attributes for each object in the input image can be used to construct characterisations that the AA-CBR can reason with. Attributes may include the shape, colour, size, material, and position of each object in the image. Labelled data points can be passed through the SA component, characterised and used to construct a casebase that can be used to construct an AF that makes predictions for unlabelled data points.
We now consider how to integrate AA-CBR with a neural component for image classification.

\paragraph{Feature Combination.} AA-CBR has no inherent ability to conduct feature weighting. Whilst the slot-attention module is useful for extracting sets of features that can be reasoned with using AA-CBR, as these two components are trained separately, we have no way of knowing whether these features are useful for the downstream classification task. To address this issue, the DEAr pipeline~\cite{DEAr} employs feature selection, discarding features deemed unimportant, e.g as determined by the weights of an autoencoder. However, discarding features may result in the removal of those necessary for classifying a specific class. This is a problem if, for example, a class depends on the existence of both a small metal cube and a small sphere whose material is irrelevant to the classification. The material of an object is important in one instance but unimportant in the other, so we cannot simply discard the attribute, but we cannot keep it for all instances either. Instead, we propose combining the individual features into super-features and then using feature selection to determine which super-features to use. For example, for an image that contains a small $(sm)$ metal $(m)$ cube $(cu)$ and a small $(sm)$ rubber $(ru)$ sphere $(sp)$, we %can represent this with 
may use two super-features: $\{sm\_m\_cu, sm\_sp\}$. By applying feature selection over super-features, we can extract only the properties necessary for a classification\footnote{This approach comes at the expense of a combinatorial number of super-features to select over. Future work should look at feature combination methods that can be learned with the model.}. So if the small sphere super-feature is selected, any time the model encounters a small rubber sphere, it will only ever be represented as simply a small sphere. The useful feature combinations can be tuned for a given AA-CBR Model. 

\paragraph{Casebase Reduction.} Furthermore, as traditional AA-CBR uses each data point as an argument, the number of arguments can be very large. This can be problematic as the outputs from the neural-component can introduce noise, if, for example, an object is incorrectly predicted in the image or there are confounding variables in the image. Additionally, for large datasets, constructing the AF with AA-CBR may be a slow process given the algorithm has a worst-case time complexity of $O(n^{3})$. Moreover, AFs with a large number of arguments are cognitively intractable, that is, they can be difficult for humans to interpret. To reduce the number of arguments, we can first apply a clustering algorithm (e.g., kMeans clustering~\cite{kMeans}) and use the cluster centroids as the data points in the casebase. We can further filter the casebase by removing centroids where the sum of the confidence scores of each predicted object does not meet a set threshold. Both these approaches filter out noisy and uncertain predictions from the neural-component whilst reducing the casebase size. The number of centroids and the value of the thresholds can be tuned as hyperparameters for an AA-CBR model.  

% For example, if an image contains a large ($l$) green ($g$) cylinder ($cy$), a blue ($b$) cylinder with an unknown size, we could just represent these two objects as a single cylinder ($cy$) feature.

\paragraph{Multi-Class Classification.} As AA-CBR computes a grounded extension which only determines if an argument is accepted or not, it is only suitable for binary classification. To apply it to a multi-class classification task, we can use a tournament-style classification strategy in which we chain various AA-CBR models together. We decided to use a One-Vs-Rest (OvR) approach 
%as depicted in Figure~\ref{fig:OvR} 
in which the AA-CBR model selects one class as the \textit{focus class} and aims to distinguish a data point between this focus class and all other classes. For those data points classified as `other', we can again select a new focus class and deploy the same strategy with the focus class excluded. We can continuously apply this until the final two classes remain, in which case the final model simply distinguishes between these two classes. We can tune the order of application of the focus classes as well as whether each model uses the focus class or `other' as the default outcome to optimise model performance. 

%\begin{figure}[h]
%\centering
%\resizebox{0.90\textwidth}{!}{
%\begin{tikzpicture}[
%    node distance=3cm,
%    every node/.style={font=\large},
%    model/.style={ellipse, draw, minimum width=3cm, minimum height=1.2cm},
%    arrow/.style={-{Latex}, thick},
%    lbl/.style={font=\large}
%  ]
%  % Main models in a row
%  \node[model]        (M1)              {AA-CBR Model 1};
%  \node[model,right=of M1] (M2)         {AA-CBR Model 2};
%  \node[model,right=of M2] (M3)         {AA-CBR Model 3};
%
%  % "match" outputs above each model
%  \node[lbl, below=1cm of M1] (O0) {0};
%  \node[lbl, below=1cm of M2] (O1) {1};
%  \node[lbl, below=1cm of M3] (O2) {2};
%
%  % final "else" output at right of M3
%  \node[lbl, right=2cm of M3] (O3) {3};
%
%  % match‐arrows (downwards)
%  \draw[arrow] (M1.south) -- (O0);
%  \draw[arrow] (M2.south) -- (O1);
%  \draw[arrow] (M3.south) -- (O2);
%
%  % not‐arrows (horizontal)
%  \draw[arrow] (M1.east) -- node[above] {Not 0} (M2.west);
%  \draw[arrow] (M2.east) -- node[above] {Not 1} (M3.west);
%  \draw[arrow] (M3.east) -- (O3);
%\end{tikzpicture}
%}
%\caption{Graphical depiction of the One-Vs-Rest (OvR) strategy for %extending 
%applying AA-CBR to multi-class classification.}
%\label{fig:OvR}
%\end{figure}

\paragraph{Supported AA-CBR.} We also experiment with Supported AA-CBR ~\cite{supported-aacbr}, in which we modify the AA-CBR model to additionally include supporting relationships between arguments with the same label. Supports allow for more arguments to be included in the classification process, as the AA-CBR minimality condition can otherwise result in excluded cases~\cite{monotonicity-and-noise-tolerance}. The support relation is interpreted as an indirect attack: if argument $\argalpha$ supports argument $\argbeta$, and $\argbeta$ attacks argument $\arggamma$, then $\argalpha$ is also considered to indirectly attack $\arggamma$. We can then apply the grounded semantics with the set of attacks and indirect attacks. By including supports, we ensure the cases in the casebase are used more effectively. Each model in the tournament can treat whether to use supports or not as an additional hyperparameter.

\paragraph{Instantiating AA-CBR.} AA-CBR relies on the use of a partial order over the case characterisations to be defined. We experiment with two possible characterisation methods and corresponding partial orders. For the first characterisation approach, we let $X = \mathbb{F}$, where $\mathbb{F}$ is the set of (super-)features, and define the exceptionality partial order with the subset relation $\subseteq$. For example, consider that the slot attention module determines that an image contains a cube $(cu)$ and a large $(l)$ cylinder $(cy)$. If we know such an image belongs to class 0, it can be represented as $\argalpha = (\{cu, l\_cy\}, 0)$ and would be considered an exception to a image with just a cube, belonging to class 1: $\argbeta = (\{cu\}, 1)$. This approach is as defined in the original proposal of AA-CBR~\cite{aa-cbr}.

However, this characterisation approach may prove problematic, given that an image may contain multiple instances of the same object with the same properties. %If we use a simple set-based characterisation, we cannot include this fact. 
Instead, we propose a novel count-based characterisation. Each case is therefore characterised as $(f_{i}:n_{\argalpha, i} | f_{i} \in \mathbb{F})$, where $n_{\argalpha, i}$ is the number of times the feature appears in case $\argalpha$. For example, if an image in class 0 contains two cubes and a single large cylinder, it would be represented as $((cu:2, l\_cy: 1), 0)$\footnote{We chose to exclude features with a count of 0 from the notation. So in this case, we know implicitly that this image has, for example, 0 spheres.}. We define exceptionalism as $(f_{i}: n_{\argalpha, i} | f_{i} \in \mathbb{F}) \prec (f_{i}: n_{\argbeta, i} | f_{i} \in \mathbb{F})$ iff $\forall i \in [0, m], n_{\argalpha, i} \leq n_{\argbeta, i}$ and $\exists i \in [0, m],  n_{\argalpha, i} < n_{\argbeta, i}$ where $m$ is the total number of features. For example, a class 0 image with two cubes and a single large cylinder would be considered an exception to an image in class 1 with just two cubes represented as $((cu: 2), 1)$ because the image in class 0 has at least as many cubes as the image in class 1 but has more large cylinders. We can treat the choice of partial order as a hyperparameter for each AA-CBR in the tournament, swapping between different characterisations and partial orders as necessary for the classification task.

\begin{wrapfigure}[11]{r}{0.4\textwidth}
\centering
        \resizebox{0.4\textwidth}{!}{
        \hspace{-1em}
        \begin{tikzpicture}[main/.style = {draw=none, font=\large}]
    
            \node[main] (C0) at (0, 0)  {\small $C_{0}: (\emptyset, 0)$};
            \node[main] (C1) at (-2, 1) {\small $C_{1}: (\{l\_cy\}, - )$};

            \node[main] (C2) at (0, 3)  {\small $C_{2}: (\{l\_cy, l\_cu, l\_cy, s\_sp\}, - )$};
            \node[main] (C3) at (-2, 2)  {\small $C_{3}: (\{l\_cy, l\_cu\}, 0)$};
            \node[main] (C4) at (2, 1)  {\small $C_{4}: (\{s\_sp\}, - )$};
            
            \node[main] (N) at (2, 2)  {\small $C_{N}: (\{l\_cy, l\_cu, cu\}, ?)$};

            \draw[-{Latex[length=1.5mm, width=2mm]}] (C1) -- (C0);                   
            \draw[-{Latex[length=1.5mm, width=2mm]}] (C3) -- (C1);                   
            \draw[-{Latex[length=1.5mm, width=2mm]}] (C2) -- (C3);              
            \draw[-{Latex[length=1.5mm, width=2mm]}] (C4) -- (C0);              
            \draw[draw=red,-{Latex[length=1.5mm, width=2mm]}] (N) -- (C2);              
            \draw[draw=red,-{Latex[length=1.5mm, width=2mm]}] (N) -- (C4);              

%            \node[main] (K) at (5, 1) {Key};
%            \node[main] (a1) at (4, 0.5) {};
%            \node[main] (a2) at (5.5, 0.5) {Attacks};
%            \draw[-{Latex[length=1.5mm, width=2mm]}] (a1) -- (a2);              
%            \node[main] (i1) at (4, 0) {};
%            \node[main] (i2) at (6.5, 0) {Irrelevance Attacks};
%            \draw[draw=red,-{Latex[length=1.5mm, width=2mm]}] (i1) -- (i2);              
            
            \begin{scope}[transparency group, opacity=0.5]
            \end{scope}
    
        \end{tikzpicture} 
        }
    \caption{An example AA-CBR model.}
    \label{fig:example-1}
\end{wrapfigure}
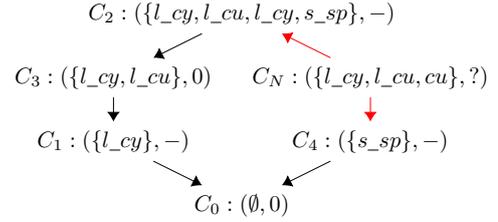

Figure \ref{fig:example-1} shows a small example AA-CBR model with just four cases in the casebase using the first characterisation approach \footnote{The real models contain the number of cluster centres in their casebase.}. The new case, $C_N$ contains a large cylinder $(l\_cy)$, a large cube $(l\_cu)$ and another cube ($cu$) but no small sphere $(s\_sp)$. Cases in the casebase attack all those with a different outcome and a subset of features, subject to the minimality condition (black arrows). Under the grounded semantics, case $C_0$ is accepted, so $C_N$ is predicted to be in class 0. This can be explained as follows: the debate starts with $C_0$, the default rule assigning class 0. Case $C_1$ challenges this, since it has a large cylinder but belongs to another class. $C_3$ then defeats $C_1$, as it includes both a large cylinder and a large cube but belongs to class 0. $C_2$ also tries to argue for the other class, but it is dismissed since it contains a small sphere, making it irrelevant to $C_N$. Similarly, $C_4$ is dismissed for the same reason.

\section{Experiments}
% \todoin{
% \begin{itemize}
%     \itemdone Explain the CLEVR-Hans dataset(s)
%     \itemdone Discuss the baseline models
%     \itemdone Explain how hyperparameters were tuned (random search)
%     \itemdone provide the table of hyperparameters
%     \itemdone CLEVR-Hans3/6 results
%     \itemdone Results discussion
%     \itemdone refer to ablation study in the appendix
%     \itemtodo Approximate AA-CBR? If we have space
% \end{itemize}
% }

\paragraph{Datasets:} We evaluate our approach on the CLEVR-Hans3 and (a modified variant of) the CLEVR-Hans7 datasets~\cite{RefWorks:RefID:2-stammer2021concept:}. The CLEVR-Hans datasets contain images with objects of varying shapes, sizes, materials and colours grouped into three and seven classes respectively. These classes are constructed from rules about the objects; for example, any image that contains a large cube and a large cylinder is in the first class.
%Note that the colour, position and materials of these objects and the existence of other objects in the image are irrelevant for the rule defining this first class. 
However, CLEVR-Hans has the additional difficulty of including confounding variables, which are designed to assess if the system can ideally learn the most general rules about a class, or instead learns more specific rules based on its training data. For example, in the training/validation sets of CLEVR-Hans, all large cubes in the images of the first class are grey, but the rule itself is not dependent on the colour of the cube. Meanwhile, the test set contains images labelled as the first class with large cubes and large cylinders of any colour.

CLEVR-Hans7 introduces four additional classes to CLEVR-Hans3, which include more complex rules that consider the number of objects, the absolute position of objects (e.g. left/right side of the image) and the relative position of the objects (e.g. is object A in front of object B). However, with our feature extraction method, relative positioning would lead to a combinatorial number of features with respect to the number of objects in the image. For this reason, we chose to exclude the one class from the dataset that relies on relative positioning in its class rule for our testing. We refer to our modified dataset as `CLEVR-Hans7 (modified)'.    

\paragraph{Hyperparameter Tuning} The integrations of AA-CBR into our SAA-CBR pipeline result in several possible parameters to tune. This includes the order of classes in which we apply the OvR strategy. Additionally, for each AA-CBR model, we must tune which class is the default, how cases are characterised, the number of representative samples used to build the AF, the threshold value, and whether the model uses supports. Due to the large number of parameters, we employed a random search to identify which could lead to the best performance. Selected hyperparameters are specified in Appendix~\ref{appendix:hyperparameter-tuning}.

\paragraph{Baseline models} We compare our proposed SAA-CBR method against four baseline models. The first is a ResNet18~\cite{restnet}, which is a deep neural network consisting of convolutional layers and residual (skip) connections. We selected this model as it offers a simple baseline for a neural-based approach that makes no use of symbolic reasoning to resolve the classification. The model was trained for 100 epochs using the Adam Optimiser~\cite{ADAM-optimiser} with a learning rate of 0.0001. 
The second baseline selected consists of a Slot Attention module in a pipeline with an MLP handling the final classification. This allows us to demonstrate whether the MLP is capable of learning the reasoning mechanism that a symbolic module is capable of. We train this in a pipeline fashion, with the SA module trained separately from the MLP to enable direct comparison with our approach. The MLP consists of a three-layer network with ReLU activations for the hidden layers and a softmax activation for the final layer. The MLP was trained for 100 epochs with the Adam optimiser with a learning rate of 0.001. 
We used OC-NAL~\cite{jacob2025objectcentric} for the third baseline model for a direct comparison of a pipeline using an SA module followed by an argumentative reasoner. This pipeline uses Assumption-Based Argumentation (ABA)~\cite{ABA-1,ABA-2} for the symbolic reasoning. This is a form of structured argumentation, corresponding in OC-NAL to logic programs with negation as failure.
% in which the semantics of a logic program can be resolved with argumentation. The logic program and argumentation semantics can be used to make classifications
The %logic program 
ABA framework is itself learned from positive and negative examples by iteratively applying rules from the ABALearn algorithm~\cite{RefWorks:RefID:24-de2023aba}.
Finally, we compare our approach to NS-CL~\cite{RefWorks:RefID:2-stammer2021concept:}, to compare against a neurosymbolic model that does not make use of argumentation and is trained end-to-end. We used the pre-trained weights for the SA module for NS-CL and the set transformer trained with an Adam optimiser, with a learning rate of 0.0001 for 50 epochs.

\subsection{Evaluation}

\begin{table}[h]
    \centering
    \begin{tabular}{|l||c|c||c|c|}
        \hline
        \multicolumn{5}{|c|}{\textbf{Validation Set}} \\
        \hline
        \multirow{2}{*}{\textbf{Model}} & \multicolumn{2}{c||}{\textbf{CLEVR-Hans3}} & \multicolumn{2}{c|}{\textbf{CLEVR-Hans7 (modified)}} \\
        \cline{2-5} & Accuracy & F1 & Accuracy & F1 \\
        \hline
         ResNet18 & $59.02 \pm 0.00 $  & $59.04 \pm 0.00$ & $30.96 \pm 0.00$ & $31.09 \pm 0.00$ \\
         \hline
         Slot Attention w/ MLP & $55.00 \pm 0.44$ & $54.36 \pm 0.44$ & $\underline{62.90 \pm 0.20}$  & $61.24 \pm 0.25$ \\
         \hline
         OC-NAL & $66.87 \pm 0.32$  & $65.47 \pm 0.29$ & $15.57 \pm 0.46$ & $11.62 \pm 0.43$ \\
         \hline
         NS-CL & $\mathbf{98.86 \pm 0.07}$ & $\mathbf{ 98.86 \pm 0.07}$ & $\mathbf{96.00 \pm 0.18}$ & $\mathbf{96.00 \pm 0.18}$ \\
         \hline
         SAA-CBR (Ours) & $\underline{75.59 \pm 2.23}$ & $\underline{77.22 \pm 2.42}$ & $62.16 \pm 2.81$ & $\underline{62.56 \pm 3.00}$ \\
         \hline
    \end{tabular}
    \caption{Model performances on the CLEVR-Hans3 and CLEVR-Hans7 (modified) validation sets. The best-performing model is shown in \textbf{bold}, and the second best is \underline{underlined}.}

    \label{clevrhans_val_results}
\end{table}

\begin{table}[h]
    \centering
    \begin{tabular}{|l||c|c||c|c|}
        \hline
        \multicolumn{5}{|c|}{\textbf{Test Set}} \\
        \hline
        \multirow{2}{*}{\textbf{Model}} & \multicolumn{2}{c||}{\textbf{CLEVR-Hans3}} & \multicolumn{2}{c|}{\textbf{CLEVR-Hans7 (modified)}} \\
        \cline{2-5} & Accuracy & F1 & Accuracy & F1 \\
        \hline
         ResNet18 & $48.71 \pm 0.00 $  & $47.13 \pm 0.00$ & $29.44 \pm 0.00$ & $29.57 \pm 0.00$ \\
         \hline
         Slot Attention w/ MLP & $52.97 \pm 0.77$ & $52.22 \pm 0.78$ & $58.33 \pm 0.57$  & $56.37 \pm 0.56$ \\
         \hline
         OC-NAL & $68.76 \pm 0.41$  & $67.57 \pm 0.40$ & $15.75 \pm 0.72$ & $11.79 \pm 0.54$ \\
         \hline
         NS-CL & $\mathbf{82.43 \pm 0.12}$ & $\mathbf{81.42 \pm 0.13}$ & $\mathbf{91.29 \pm 0.12}$ & $\mathbf{91.22 \pm 0.12}$ \\
         \hline
         SAA-CBR (Ours) & $\underline{75.13 \pm 2.16}$ & $\underline{74.77 \pm 2.28}$ & $\underline{62.87 \pm 2.81}$ & $\underline{63.19 \pm 3.00}$ \\
         \hline
    \end{tabular}
    \caption{Model performances on the CLEVR-Hans3 and CLEVR-Hans7 (modified) test sets. The best-performing model is shown in \textbf{bold}, and the second best is \underline{underlined}.}

    \label{clevrhans_test_results}
\end{table}

Table~\ref{clevrhans_test_results} showcases the macro-averaged accuracy and F1-score evaluated on the test set of both datasets across five runs. We see that %of the baseline models, 
SAA-CBR significantly outperforms all baselines except the NS-CL. We see that purely neural approaches cannot perform well on the CLEVR-Hans classification task, with the ResNet18 achieving an accuracies of $48.71\% \pm 0.00$ and $29.44\% \pm 0.00$ and the Slot Attention with MLP achieving $52.97\% \pm 0.77$ and $58.33\% \pm 0.57$ on CLEVER-Hans3 and CLEVER-Hans7 (modified), respectively. In particular, these models perform better on the validation sets, as shown in Table~\ref{clevrhans_val_results}, than on the test sets, suggesting that these models %learned concepts 
fail to generalise beyond the confounding variables. 

A form of symbolic learning reasoning appears necessary to achieve higher performance on this task without large model scaling. The introduction of argumentative reasoning with OC-NAL demonstrates how performance can be improved for CLEVR-Hans3, achieving an accuracy of $66.87\% \pm 0.32$. However, this approach fails to generalise to CLEVR-Hans7 (modified), with an accuracy of $15.57\% \pm 0.46$. It is of note that the performance of OC-NAL increased between the validation set to the test set, suggesting that the introduction of argumentative reasoning allows for this pipeline to learn rules that generalise beyond the confounders. One of the key limitations of learning %assumption-based argumentation 
ABA frameworks is ABALearn's inability to learn from a large number of examples, wherein only 10 representative samples per class could be used to learn the framework. On the other hand, the largest of the AA-CBR models in the SAA-CBR pipeline was capable of using over 900 representative samples in its learning process.
As a result, the SAA-CBR pipeline performs considerably better, achieving accuracies of $75.13\% \pm 2.16$ and $62.87\% \pm 2.81$ on CLEVR-Hans3 and CLEVER-Hans7 (modified), respectively. Notably, the average performance change between the SAA-CBR pipeline on the validation set compared to the test set is the smallest compared to any other model, with only a small decrease in accuracy for CELVR-Hans3 and an increase in accuracy for CLEVR-Hans7 (modified). As with OC-NAL, this suggests that one of the key benefits of applying argumentative reasoning here is the ability to learn a more general set of rules for classifying the data. { However, the variance for SAA-CBR is greater, which we attribute to the randomisation resulting from the initialisation of the k-Means centroids in the casebase clustering.}

The NS-CL appears to have greater learning capacity, and the set transformers are capable of learning from the full training set compared to using a subset of representative samples with AA-CBR. However, the NS-CL model exhibits a decrease in performance on the test set compared to the validation set for both CLEVR-Hans datasets (although not as pronounced for CLEVR-Hans7 (modified)), suggesting that, despite its greater learning capacity, this model struggles to generalise beyond the confounding variables. Our modifications to AA-CBR are key for our pipeline to perform as it does, demonstrated by our ablation study in Appendix~\ref{appendix:ablation-study}. NS-CL is trained end-to-end, rather than in a pipeline fashion with separate training for the neural and symbolic components, which may explain its improved performance. Future work can explore end-to-end training of both AA-CBR and slot attention.

\section{Conclusion}
% \todoin{
% \begin{itemize}
%     \item One of the first analysis of AA-CBR applied to images
%    \item First integration of AA-CBR with object-centric learning
%    \item 
% \end{itemize}
% Future Work:
% \begin{itemize}
%     \itemdone Improve case characterisation to be learned so there is no combinatorial explosion of features
%     \itemdone some improvement to AA-CBR so it can do multi-class classification directly rather than tournament style
%     \itemdone Exploration of increasing AA-CBR model capacity for learning or try scaling on larger hardware so we can see if it is capable of better performance without the need for representative samples
%     \itemdone End-to-End training of the models (or applying RL to update the SA with rewards from the AA-CBR model)
%     \itemdone Exploration of Supported AA-CBR, Preference-Based AA-CBR, Cumulative AA-CBR or Gradual AA-CBR on this task
% \end{itemize}
% }

We introduce Slot-Attention Argumentation for Case-Based Reasoning, a novel neuro-argumentative framework integrating SA and AA-CBR for image classification. We showcase how the model outperforms neural-based baseline models and other proposed methods of combining SA and (structured) argumentation. Our novel extensions to AA-CBR, namely the introduction of a One-Vs-Rest model for multi-class classification, the application of Supported AA-CBR, the use of clustering and uncertainty filtering for casebase reduction, feature combination methods and the introduction of count-based partial orders, improve the pipeline, allowing for a successful integration of symbolic argumentative reasoning via case-based reasoning with slot attention.

Further research into the combination of SA and AA-CBR should investigate how the integration between these two components can be further enhanced. For example, instead of using a feature combination method and applying feature selection after the fact, research should explore how to directly learn which feature combinations to use. Additionally, improvements to constructing representative samples that consider the downstream argumentation process may prove more effective than using k-means clustering. Such an approach may also prove to scale better, allowing for fewer necessary representative samples that can be learned from a larger dataset. Additionally, other AA-CBR variants such as Preference-Based AA-CBR~\cite{preference-based-aacbr}, Cumulative AA-CBR~\cite{monotonicity-and-noise-tolerance}, or Gradual AA-CBR~\cite{Gradual-AACBR} could be experimented with on this task. This latter model, which scores argument acceptability with a value between 0 and 1 rather than existence in an extension, may prove useful for improving multi-class classification beyond the One-Vs-Rest strategy and allow for end-to-end learning.

\begin{acknowledgments}
Gould was supported by UK Research and Innovation [UKRI Centre for Doctoral Training in AI for Healthcare grant number EP/S023283/1].
Kori was supported by UKRI through the CDT in Safe and Trusted Artificial Intelligence and  acknowledges support from the EPSRC Doctoral
Prize. Toni was partially funded by the ERC under the EU’s Horizon 2020 research and innovation programme (grant agreement No. 101020934). Toni was also partially funded by J.P. Morgan and the Royal Academy of
Engineering, UK, under the Research Chairs and Senior Research Fellowships scheme.
\end{acknowledgments}

%%
%% Define the bibliography file to be used
\bibliography{references}

\appendix

\section{Hyperparameters}
\label{appendix:hyperparameter-tuning}

\begin{table}[h]
    \centering
    \begin{tabular}{|l|c|c|}
        \hline
        \multirow{2}{*}{\textbf{Hyperparameters}} & \multicolumn{2}{c|}{\textbf{AA-CBR Models}} \\
        \cline{2-3} & Model 1 & Model 2 \\
        \hline
        Characterisation & Count & Set \\
        \hline
        Model Classifies & Class 2 vs. rest & Class 1 vs. Class 0 \\
        \hline
        Default Outcome & Rest & Class 1 \\
        \hline
        No. K-means Centroids & 900 & 500 \\
        \hline
        Threshold & \xmark & 0.9 \\
        \hline
        Uses Supports & \xmark & \cmark \\
        \hline
    \end{tabular}
    \caption{SAA-CBR argumentative component's hyperparameters for CLEVR-Hans3 dataset}
    \label{clevrhans3_hyperparams}
\end{table}

\begin{table}[h]
    \centering
    \begin{tabular}{|l|c|c|c|}
        \hline
        \multirow{2}{*}{\textbf{Hyperparameters}} & \multicolumn{3}{c|}{\textbf{AA-CBR Models}} \\
        \cline{2-4} & Model 1 & Model 2 & Model 3 \\
        \hline
        Characterisation & Count & Count & Count \\
        \hline
        Model Classifies & Class 0 vs. rest & Class 6 vs. rest & Class 3 vs. rest \\
        \hline
        Default Outcome & Rest & Rest & Rest \\
        \hline
        No. K-means Centroids & 300 & 500 & 300 \\
        \hline
        Threshold & 0.75 & 0.7 & 0.75 \\
        \hline
        Uses Supports & \xmark & \xmark & \xmark \\
        \hline
        \multicolumn{4}{|c|}{} \\
        \hline
        \multirow{2}{*}{\textbf{Hyperparameters}} & \multicolumn{3}{c|}{\textbf{AA-CBR Models}} \\
        \cline{2-4} & Model 4 & Model 5 & \\
        \hline
        Characterisation & Count & Position Count & \\
        \hline
        Model Classifies & Class 1 vs. rest & Class 4 vs. Class 5 & \\
        \hline
        Default Outcome & Class 1 & Class 4 & \\
        \hline
        No. K-means Centroids & 300 & 300 & \\
        \hline
        Threshold & 0.85 & 0.75 & \\
        \hline
        Uses Supports & \xmark & \xmark & \\
        \hline
    \end{tabular}
    \caption{SAA-CBR argumentative component's hyperparameters for CLEVR-Hans7 (modified) dataset}
    \label{clevrhans6_hyperparams}
\end{table}

Note that the difference between the Count characterisation and the Position Count characterisation is that Position Count also considers whether an object is on the left or right side of the image as part of the feature combination process, whereas the Count characterisation does not. Note that only Class 5 in CLEVR-Hans 7 (modified) uses the absolute position of the object, so it makes sense that our hyperparameter tuning would find the best performance is to only use the Position Count partial order when classifying this class.  

\section{Ablation Study}
\label{appendix:ablation-study}

We conducted an ablation study with the CLEVR-Hans3 dataset to assess which aspects of our integrations to the SAA-CBR pipeline led to improved model performance. We choose to explore how the model would perform without using a feature combination approach, without uncertainty thresholding to reduce the casebase size and without using the support relations. Note that we chose not to include the k-means clustering from the ablation study, as using the full casebase for training the AA-CBR models was infeasible given the $O(n^{3})$ time complexity. 

\begin{table}[h]
    \centering
    \begin{tabular}{|m{6cm}|c|c|c|c|}
        \hline
        \multirow{2}{*}{\textbf{Methods}} & \multicolumn{4}{c|}{\textbf{Test Set}} \\
        \cline{2-5} & Accuracy & Precision & Recall & F1 \\
        \hline
        SAA-CBR & $75.13 \pm 2.16$ & $75.31 \pm 2.10$ & $75.13 \pm 2.16$ & $74.77 \pm 2.28$ \\
        \hline
        w/o feature combination & $54.68 \pm 3.69$ & $62.97 \pm 1.53$ & $54.68 \pm 3.69$ & $47.11 \pm 3.92$ \\
        \hline
        w/o thresholding & $68.04 \pm 4.61$ & $70.70 \pm 6.00$ & $68.04 \pm 4.61$ & $66.02 \pm 4.76$ \\
        \hline
        w/o supports & $74.30 \pm 2.96$ & $74.79 \pm 2.39$ & $74.30 \pm 2.96$ & $73.73 \pm 3.44$ \\
        \hline
        w/o supports, thresholding, feature combination & $55.16 \pm 3.81$ & $62.37 \pm 2.12$ & $55.16 \pm 3.81$ & $48.88 \pm 4.76$ \\
        \hline
    \end{tabular}
    \caption{Ablation study of the SAA-CBR model on the CLEVR-Hans3 test set}
    \label{tab:ablation3test}
\end{table}

Table~\ref{tab:ablation3test} shows our results. From our ablation study on CLEVR-Hans3, we observe that the largest increase in model performance was a result of the feature combination strategy, without which the addition of supports or thresholding surprisingly reduces model performance. However, when used in combination, we see that both the thresholding and the support relation are necessary for improving the model's classification performance. Interestingly, the model's variance is considerably lower when all of these integration methods are applied, suggesting that the training variability is also dependent on making use of these integrations, even though the variability only comes from the initial weights of the Slot Attention module and the initial position of the k-means cluster centroids. 

\end{document}